\documentclass[letterpaper]{article} 
\usepackage{aaai23}  
\usepackage{times}  
\usepackage{helvet}  
\usepackage{courier}  
\usepackage[hyphens]{url}  
\usepackage{graphicx} 
\usepackage{natbib}  
\usepackage{caption} 
\usepackage{algorithm}
\usepackage{algorithmic}
\usepackage{amsmath}
\usepackage{booktabs}
\usepackage{multirow}
\usepackage{romannum}

\usepackage{amssymb}
\usepackage{pifont}
\newcommand{\cmark}{\ding{51}}%
\newcommand{\xmark}{\ding{55}}%

\usepackage{newfloat}
\usepackage{listings}
\DeclareCaptionStyle{ruled}{labelfont=normalfont,labelsep=colon,strut=off} 
\floatstyle{ruled}
\newfloat{listing}{tb}{lst}{}
\floatname{listing}{Listing}
\title{Generative Image Inpainting with Segmentation Confusion Adversarial Training and Contrastive
Learning}
\author{
    Zhiwen Zuo, Lei Zhao\thanks{Corresponding author.},
    Ailin Li, Zhizhong Wang\\
     Zhanjie Zhang, Jiafu Chen, Wei
     Xing, and
    Dongming Lu
}
\affiliations{
    College of Computer Science and Technology, Zhejiang University\\

    \{zzwcs, cszhl, liailin, endywon, cszzj, chenjiafu, wxing, ldm\}@zju.edu.cn
}

\usepackage{bibentry}

\begin{document}

\maketitle

\begin{abstract}
This paper presents a new adversarial training framework for image inpainting with segmentation
confusion adversarial training (SCAT) and contrastive learning. SCAT plays an adversarial game
between an inpainting generator and a segmentation network, which provides pixel-level local training
signals and can adapt to images with free-form holes. By combining SCAT with standard global
adversarial training, the new adversarial training framework exhibits the following three advantages
simultaneously: (1) \textit{the global consistency of the repaired image}, (2) \textit{the local fine
texture details of the repaired image}, and (3) \textit{the flexibility of handling images with free-form
holes}.  Moreover, we propose the textural and semantic contrastive learning losses to stabilize and
improve our inpainting model's training by exploiting the feature representation space of the
discriminator, in which the inpainting images are pulled closer to the ground truth images but pushed
farther from the corrupted images. The proposed contrastive losses better guide the repaired images
to move from the corrupted image data points to the real image data points in the feature representation
space, resulting in more realistic completed images. We conduct extensive experiments
on two benchmark datasets, demonstrating our model's effectiveness and superiority both qualitatively
and quantitatively.
\end{abstract}

\section{Introduction}

Image inpainting is the task that aims at estimating the missing pixels in corrupted images, which has a
wide impact on many image editing applications such as object removal, image denoising, and image
restoration, to name a few. As one of the most important research areas in computer vision, image
inpainting has drawn a significant amount of attention in the community for a long time and yet remains an
open challenge, for it is an inherently ill-posed problem. Prior to the recent dominance of deep
generative inpainting methods, early traditional inpainting
methods~\cite{barnes2009patchmatch,bertalmio2000image,bertalmio2003simultaneous,
criminisi2004region,huang2014image} typically utilize low-level features to fill holes by either
propagating surroundings to the missing regions in a diffusive manner or iteratively searching for the
best-matched patches in the internal image contexts or external image datasets. Though these
methods perform considerably well in tasks like background inpainting, they lack the understanding of
high-level semantics and can not capture the global structure of the image, thus failing to generalize
when the missing regions are large or contain unique image statistics. In contrast, recent deep
generative approaches~\cite{pathak2016context,iizuka2017globally,yu2018generative,yu2019free},
trained on large-scale datasets, benefit from the rich hierarchical features learned by the deep
convolutional neural networks
(DCNNs)~\cite{krizhevsky2012imagenet,simonyan2014very,szegedy2015going,he2016deep} and
the powerful generative capacity of generative adversarial networks
(GANs)~\cite{goodfellow2014generative}, capable of synthesizing semantically correct and visually
plausible inpainting results.

\begin{figure}[t]
	\centering
	\includegraphics[width=\columnwidth]{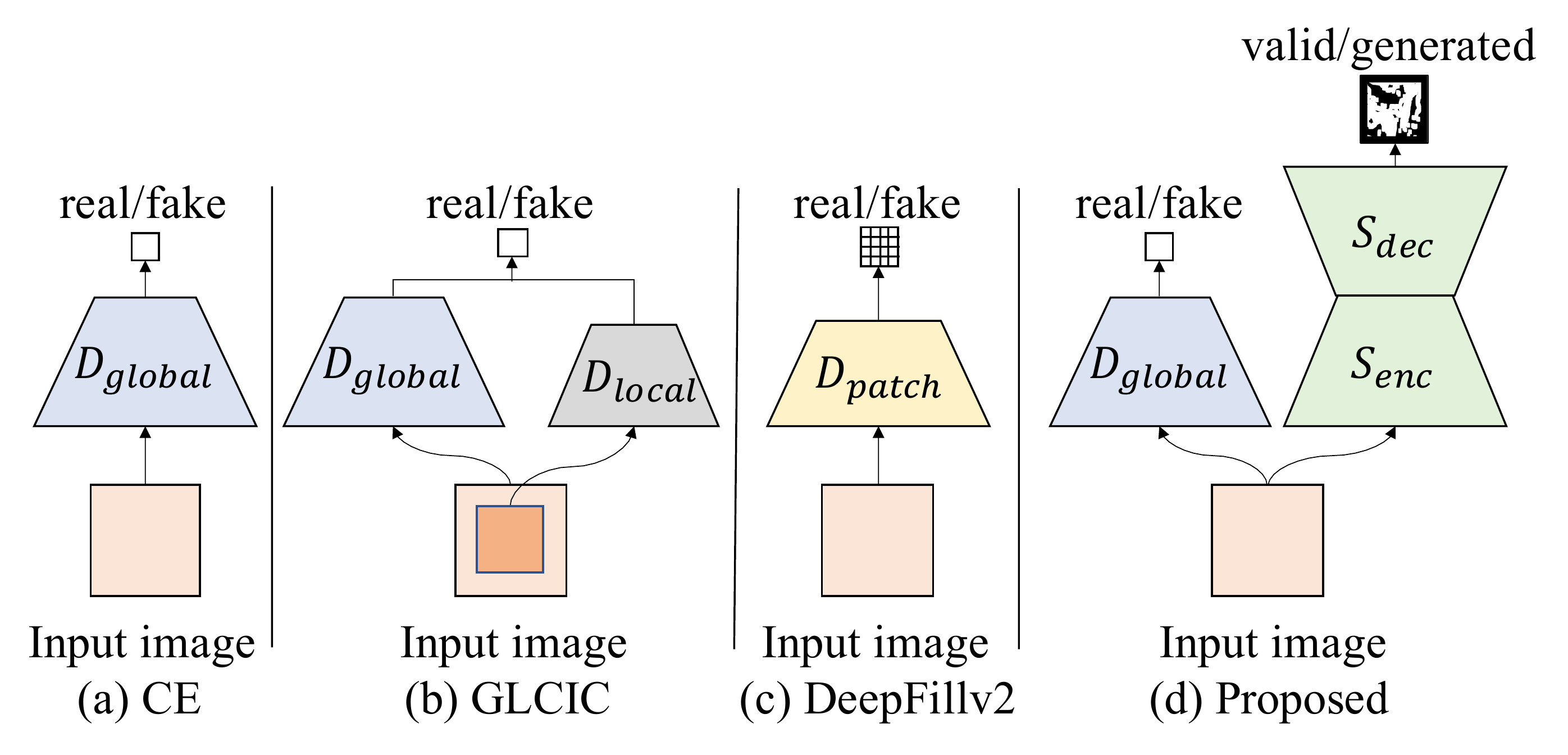}
	\caption{Different adversarial training frameworks of GAN-based inpainting models. In addition to a
	global discriminator, we introduce a segmentation netowrk $S$ in our model, which tries to label the
	valid and the generated regions in the input image.}
	\label{fig:adv_comp}
\end{figure}

Among the deep generative inpainting methods, Context Encoder (CE)~\cite{pathak2016context} first
proposed to train an encoder-decoder inpainting network with a combination of a reconstruction loss
and an adversarial loss, showing the significance of adversarial training in bringing sharper predictions.
Since then, exploring and designing more advanced and sophisticated adversarial training frameworks
have become an important research direction for improving GAN-based inpainting models. For
example, as CE only used a global context discriminator to distinguish real images from completed ones
as a whole, GLCIC~\cite{iizuka2017globally} proposed to use global and local context discriminators
simultaneously, which provides both global coherent consistency and local fine texture details for the
inpainting results. However, the local discriminator of GLCIC can only handle pre-defined fixed-sized
holes, making it inflexible to process images with arbitrary free-form holes. In order to solve the
problem, DeepFillv2~\cite{yu2019free} proposed to adopt PatchGAN from the literature on
image-to-image translation~\cite{isola2017image}. The PatchGAN discriminator only consists of a few
convolutional layers, which distinguishes patches of ground truth images from those of the repaired
images. Moreover, \citet{yu2019free} applied spectral normalization~\cite{miyato2018spectral} to
further stabilize PatchGAN training. Thenceforth, SN-PatchGAN has been vastly used in recent
GAN-based inpainting models to handle images with free-form holes. However, DeepFillv2 does not
involve a global discriminator for training, which may fail to guarantee the global consistency of the
repaired images. We show the adversarial training frameworks of existing GAN-based inpainting
models in Figure \ref{fig:adv_comp}. Additionally, the comparison of these frameworks is summarized
in Table \ref{tab:adv_comp}, where the comparison is based on three dimensions: (1) the global
consistency of the repaired image, (2) the local fine texture details of the repaired image, and (3) the
flexibility of handling images with arbitrary free-form holes. Clearly, none of the existing adversarial
training frameworks can satisfy the three requirements at the same time.

Motivated by the analysis above, we propose a new adversarial training framework for image inpainting
by first introducing a novel segmentation confusion adversarial training (SCAT) paradigm. The
proposed SCAT is inspired by \textit{how humans recognize a low-quality repaired image}. We human
beings can easily judge whether a repaired image is of poor quality by looking for the regions that are
distorted or inconsistent with the surrounding textures in the image. To mimic such human behavior,
we introduce a segmentation network to label the generated and the valid regions in the input
images, which is essentially a two-class dense semantic segmentation task. On the contrary, the
inpainting generator tries to deceive the segmentation network by filling the missing regions with more
visually plausible and consistent contents, making it more difficult for the segmentation network to label
the two regions. We further combine the introduced SCAT with the global per-image adversarial
training (see Figure \ref{fig:adv_comp}(d)),  where the final framework can meet the three
aforementioned requirements simultaneously. Complementary to the global per-image adversarial
training, the SCAT additionally provides fine-grained pixel-level training signals and the flexibility of
handling images with arbitrary free-form holes for our framework, significantly improving the image
quality and visual consistency of the inpainting results.

On the other hand, training GAN is known to be notoriously difficult, for it requires searching a Nash
equilibrium of a non-convex game in an extremely high-dimensional parameter
space~\cite{goodfellow2016nips,salimans2016improved}. Besides, the discriminator has to adapt to
the continuously changed generated distribution during the training procedure to perform the
classification task, which means GAN's training is in a non-stationary
environment~\cite{salimans2016improved,thanh2020catastrophic}.
Consequently, GAN-based inpainting models can exhibit unstable and cyclic issues, which may cause
degraded inpainting results. To stabilize and improve our model's training, we further propose
contrastive learning losses by exploiting the feature representation space of the discriminator, in which
the inpainting images are constantly pulled closer to the ground truth images but pushed farther from
the corrupted images. As the training process of image inpainting can be regarded as learning a
mapping from the corrupted images to the ground truth images, our proposed
contrastive losses can better guide the process with their pull and push forces, which brings more
realistic inpainting results. To the best of our knowledge, this is the first effective usage of contrastive
learning in image inpainting.

We conduct extensive experiments on two benchmark datasets: Places2~\cite{zhou2017places} and
CelebA~\cite{liu2018large}, demonstrating our model's effectiveness and superiority both qualitatively
and quantitatively.

The main contributions in the paper are four-fold:  \textbf{(\romannum{1})} we present a new
adversarial training framework for image inpainting with segmentation confusion adversarial training
(SCAT) and contrastive learning; \textbf{(\romannum{2})} the proposed SCAT helps provide local
pixel-level training signals and adapt to images with arbitrary free-form holes for our framework;
\textbf{(\romannum{3})} the proposed contrastive learning losses stabilize and improve our
model's training, resulting in more realistic inpainting images; and \textbf{(\romannum{4})} extensive
experiments on two benchmark datasets have been conducted to verify our model's effectiveness and
superiority.

\begin{table}[t!]
	\centering
	\resizebox{\columnwidth}{!}{
		\begin{tabular}{ccccc}
			\toprule
			Models & CE & GLCIC &
			DeepFillv2 & Proposed \\
			\midrule
			global consistency & \cmark &\cmark & \xmark & \cmark\\
			local texture details & \xmark & \cmark & \cmark & \cmark \\
			free-form holes & \xmark & \xmark & \cmark & \cmark \\
			\bottomrule
	\end{tabular}}
	\caption{Comparison of different adversarial training frameworks for image inpainting.}
	\label{tab:adv_comp}
\end{table}

\begin{figure*}[t]
	\centering
	\includegraphics[width=0.9\linewidth]{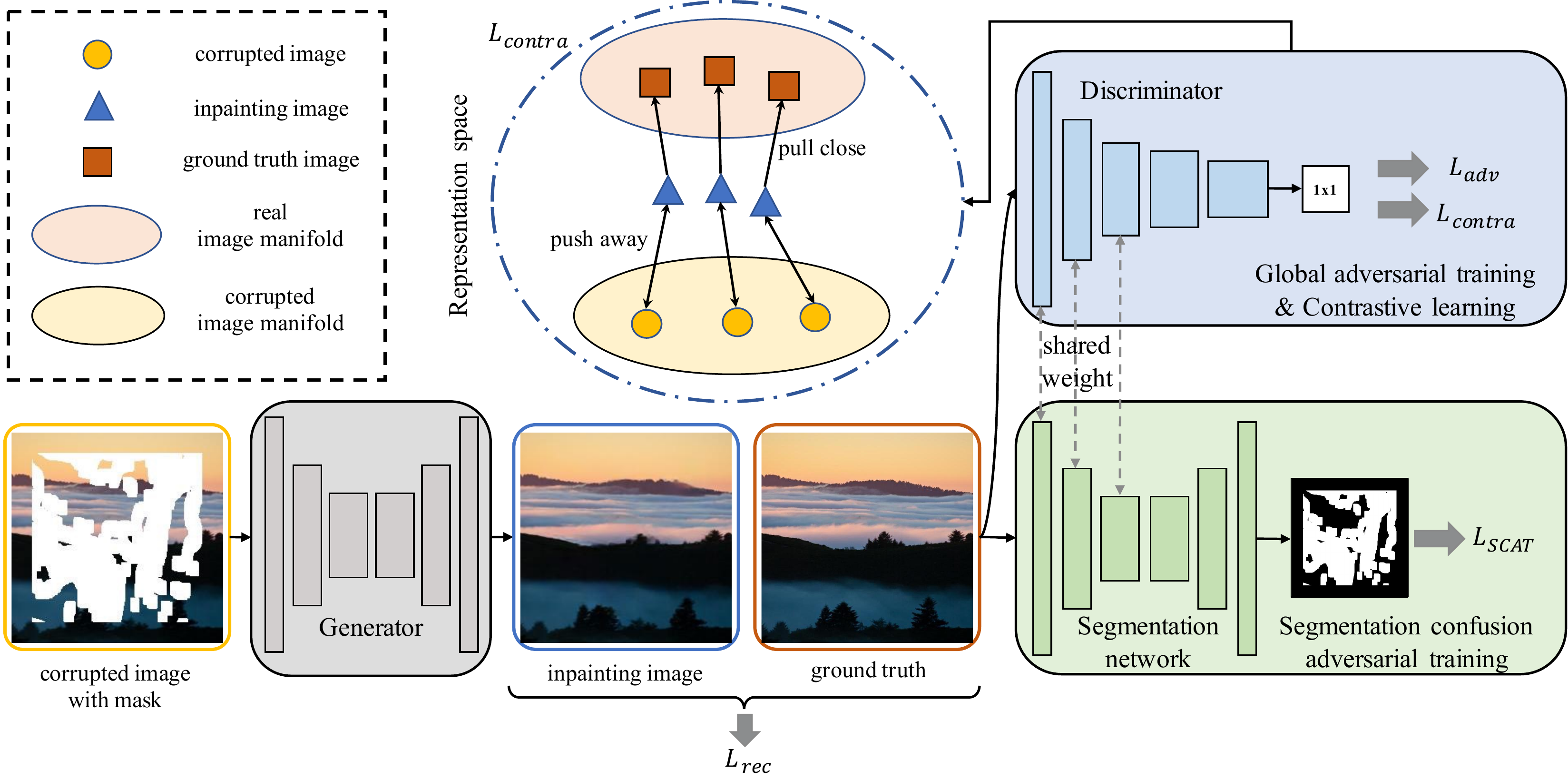}
	\caption{The overall framework of the proposed model, which consists of an inpainting generator $G$, a discriminator $D$, and a segmentation network $S$.}
	\label{fig:model}
\end{figure*}

\section{Related Works}

\subsection{Generative Adversarial Networks}
GANs are composed of two networks: a discriminator that distinguishes real data samples from
generated samples and a generator that tries to generate samples to fool the discriminator. Many
important works have been proposed to improve the original GAN for more stabilized training or
producing high-quality samples, such as by proposing better loss functions or
regularizations~\cite{arjovsky2017wasserstein,gulrajani2017improved,
	miyato2018spectral},
changing network structures~\cite{radford2015unsupervised,karras2019style,
	brock2018large,schonfeld2020u}, or combining GANs with inference networks or
	autoencoders~\cite{donahue2016adversarial,dumoulin2016adversarially,larsen2016autoencoding,
		srivastava2017veegan,ulyanov2018takes}. Our proposed SCAT is related to U-net
	GAN~\cite{schonfeld2020u}, while it differs from U-net GAN in two aspects: (1) SCAT is inspired by \textit{how humans recognize
	low-quality repaired images}; (2) In contrast to simply classifying all pixels as real or fake in U-net GAN, SCAT identifies the generated and the valid regions in the input images, which is specially tailored for image inpainting tasks.

\subsection{Deep Generative Inpainting Methods}
The seminal CE~\cite{pathak2016context} work first proposed a global adversarial training framework
for image inpainting. GLCIC~\cite{iizuka2017globally} improved it by using global and local context
discriminators simultaneously, exhibiting globally and locally consistent inpainting results. GLCIC also
proposed to use
dilated convolutional layers instead of standard ones in the feature encoder to enlarge the receptive
field of the encoded feature for better context reasoning. Further, DeepFillv1~\cite{yu2018generative} proposed the attention module in its two-stage coarse-to-fine framework, leveraging the neural patch similarities between the known and the missing regions of the inpainting image for better results. While previous works mainly focused on repairing images
with centered rectangular holes, PConv~\cite{liu2018image} proposed partial convolutions to handle
images with irregular free-form holes, where the convolution is masked and renormalized to be
conditioned only on valid pixels. DeepFillv2~\cite{yu2019free} proposed a learnable mask-update
mechanism in its framework plus SN-PatchGAN~\cite{miyato2018spectral,isola2017image} to improve free-form image
inpainting.  There are
also many methods proposing to solve the difficult hole-filling problem in a
progressive~\cite{guo2019progressive,zeng2019learning} or recurrent
way~\cite{li2020recurrent,zeng2020high}, achieving promising results. Some studies focus on
incorporating structure priors (\textit{e.g.,}
edges~\cite{nazeri2019edgeconnect,guo2021image} or frequency knowledge~\cite{yu2021wavefill,suvorov2022resolution} in their frameworks. Notably, many
methods~\cite{zheng2019pluralistic,zhao2020uctgan,zhao2021large,wan2021high} aimed at achieving diversified results for image inpainting. Very recently, transformer-based image inpainting frameworks~\cite{wan2021high,li2022mat,dong2022incremental,liu2022reduce} shined with stunning results as transformers are more expressive than DCNNs. The most related work to ours in the literature is PAL4Inpaint~\cite{zhang2022perceptual}, which trained a segmentation model to detect inpainting perceptual artifacts and apply the model for inpainting model evaluation and iterative refinement. The main distinction between ours and theirs is that we train the segmentation network and the inpainting generator adversarially.

\subsection{Contrastive Learning}
In recent years, contrastive learning has achieved great success in self-supervised representation learning~\cite{chen2020simple,he2020momentum}. Many low-level vision tasks have benefited from utilizing contrastive learning as well. For example, contrastive
learning has been utilized in one-sided unpaired image-to-image translation~\cite{park2020contrastive}, class-conditioned image generation~\cite{kang2020contragan}, artistic style transfer~\cite{chen2021artistic}, and single image dehazing~\cite{wu2021contrastive}. In image inpainting, \citet{ma2020can} proposed to improve context encoding by using a self-supervised inference network~\cite{he2020momentum}, which pulls two identical images with different masks to be close in the context feature space. We argue its strategy may be unreasonable, because the two identical images with different masks may differ significantly. Unlike previous approaches, we propose to use contrastive learning to stabilize and improve our proposed inpainting model's training by exploiting the feature representation space of the discriminator, in which the inpainting images are better guided to move from the corrupted images to the real images.

\begin{figure*}[t]
	\centering
	\includegraphics[width=\linewidth]{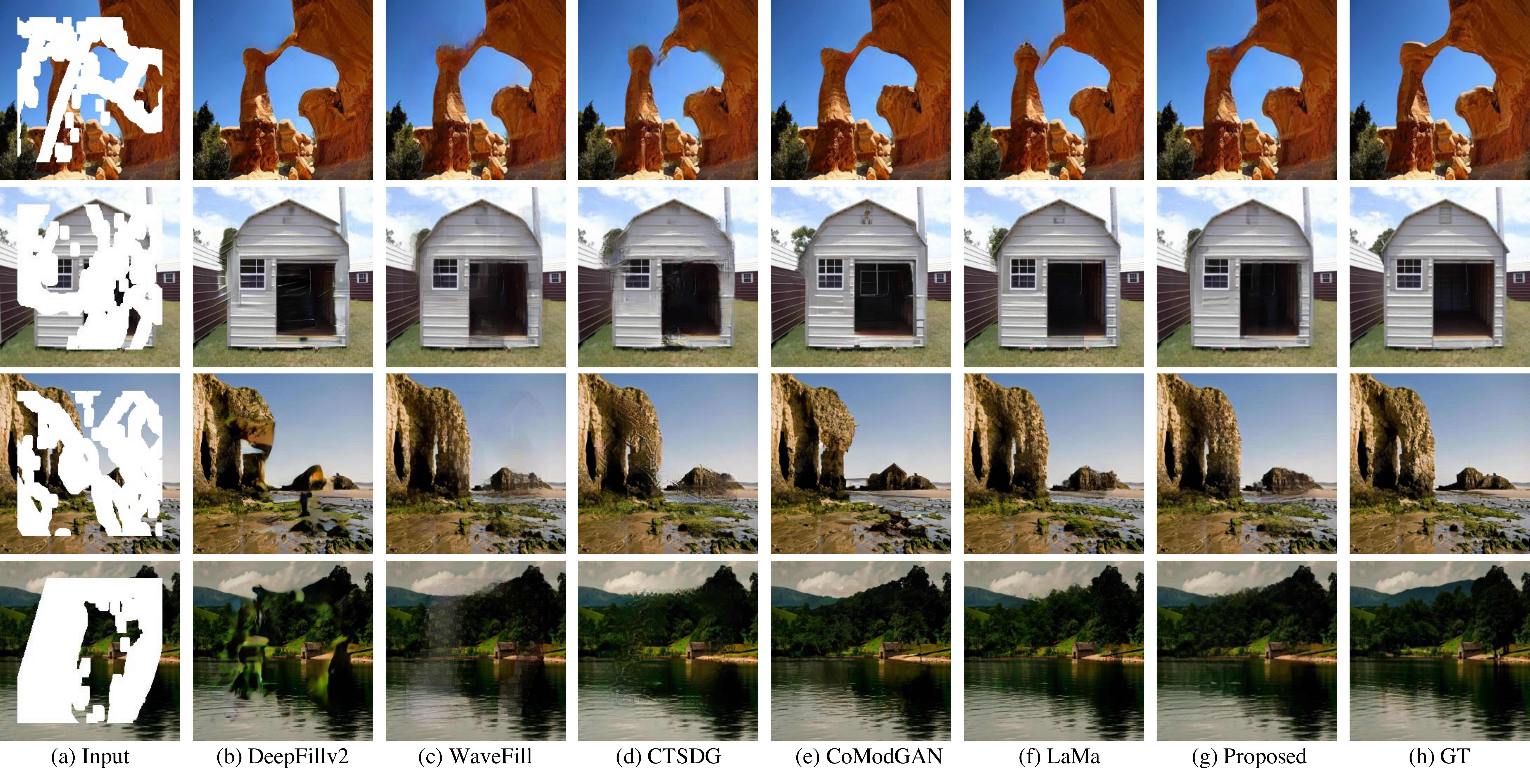}
	\caption{Qualitative results on Places2. Please zoom in to see more details.}
	\label{fig:places2}
\end{figure*}

\section{Proposed Method}
\subsection{Notations}
The overall framework of the proposed model is illustrated in Figure \ref{fig:model}, where we assume that the real images and the corrupted images lie in two manifolds which are denoted as the real image manifold and the corrupted image manifold, respectively.
During training, given an ground truth image $x\in \mathbb{R}^{H\times W \times 3}$, a mask $m\in \mathbb{R}^{H\times W \times 3}$ is applied to it to get the corrupted input image $\tilde{x}=x\odot m$, where each pixel of the mask $m$ contains values of either $0$s or $1$s, denoting the missing and the valid regions of the corrupted image, respectively. In both training and inference phases, the inpainting generator $G$ takes the corrupted input image $\tilde{x}$ and the mask $m$ as the inputs and then outputs an estimate of the repaired image $\hat{x} \in \mathbb{R}^{H\times W \times 3}$ such that $\hat{x}=G(\tilde{x},  m)$. The final inpainting result $\bar{x}$ is obtained by combining the masked regions of $\hat{x}$ and the valid regions of $\tilde{x}$ where $\bar{x} = (1-m)\odot\hat{x} + m\odot \tilde{x}$.

\subsection{Segmentation Confusion Adversarial Training}
The proposed segmentation confusion adversarial training (SCAT) plays an adversarial game between
the inpainting generator $G$ and the segmentation network $S$. Different from the global per-image
adversarial training in standard GANs, the segmentation network $S$ tries to label the valid and the generated regions in the input image at a pixel level according to the input mask image. On the contrary, the inpainting generator $G$ tries
to improve its inpainting results by filling the missing regions with more realistic and coherent contents
so that the segmentation network $S$ is not able to label the two regions anymore. The segmentation
confusion adversarial training loss $\mathcal{L}_{\text{SCAT}}$ for the segmentation network $S$ and
the inpainting generator $G$ are defined as follows, respectivly:

\begin{equation}
\small
\begin{split}
&\mathcal{L}_{\text{SCAT}}(S)=\\
&-\mathbb{E} \left[ \frac{1}{HW}\sum_{i=1}^{HW} [m_i\log
S(\bar{x})_i + (1-m_i) \log (1-S(\bar{x})_i) ]  \right.\\
&\left. + \frac{1}{HW}\sum_{i=1}^{HW}[\bar{m}_i\log
S(x)_i + (1-\bar{m}_i) \log (1-S(x)_i)] \right]
\end{split}
\end{equation}

\begin{equation}
\small
\begin{split}
&\mathcal{L}_{\text{SCAT}}(G)= \\
&-\mathbb{E} \left[ \frac{1}{HW}\sum_{i=1}^{HW} \left[\bar{m}_i\log
S\left(\bar{x}\right)_i
+ (1-\bar{m}_i) \log \left(1-S\left(\bar{x}\right)_i\right) \right]\right],
\end{split}
\end{equation}
where $\bar{m}$ is a mask filled with all $1$s and the output activation function of the segmentation network is the sigmoid function. 
When training the segmentation network $S$, we use the masks $m$ and $\bar{m}$ as the supervisions for the repaired image $\bar{x}$ and the ground truth image $x$, respectively.
On the other hand, when training the inpainting generator $G$, we use the mask $\bar{m}$ as the supervision for the repaired image $\bar{x}$, encouraging $G$ to produce better repaired images such that $S$ mistakes the generated regions in the repaired image as valid regions.

\subsection{Textural and Semantic Contrastive Learning}
Two contrastive learning losses, namely textural and semantic contrastive learning losses, are
proposed to stabilize and improve our inpainting model's adversarial training.
Since the discriminator $D$ learns discriminative features to distinguish
between the ground truth images and the completed images, its feature representation space is a
natural choice for our proposed contrastive losses.  As illustrated in Figure \ref{fig:model}, the pull
force of our contrastive losses encourage realistic completed contents by matching the feature
statistics of the repaired images to those of the ground truth images, while the push force of our
contrastive losses prevent degraded or cyclic inpainting results by keeping the repaired images to be
farther and farther away from the corrupted images in the representation space of the discriminator,
stabilizing and improving the adversarial training in our model.

As DCNNs learn hierarchical feature representations ~\cite{zeiler2014visualizing}, where the shallow and deep layers' feature maps typically contain low-level (\textit{e.g.,} local texture details) and high-level features (\textit{e.g.,} global semantic information) respectively, we design the textural and semantic contrastive learning losses by utilizing the low-level and high-level features extracted from the discriminator accordingly. 

The textural contrastive loss is defined as follows:
\begin{equation}
\mathcal{L}_{contra}^{text} = \mathbb{E} \sum_{i=1}^{N} \frac{d(D_i(\bar{x}),
	D_i(x))}{d(D_i(\bar{x}),
	D_i(\tilde{x}))},
\label{eq:3}
\end{equation}
where $d(\cdot,\cdot)$, $D_i(\cdot)$, and $N$ denote a distance metric, the discriminator $D$'s $i$-th layer's output feature map, and the number of total used shallow layers, respectively. Equation \ref{eq:3} is slightly different from the standard InfoNCE's formulation due to the relatively large dimensionality of the shallow layer's feature maps. In implementation, we adopt the $L$1-norm as the distance metric.

On the other hand, the semantic contrastive learning loss is built upon the discriminator $D$'s last output feature map as shown below:
\begin{equation}
\label{eq:4}
\resizebox{\columnwidth}{!}{$\mathcal{L}_{contra}^{sem} = -\mathbb{E} \left[ \log
	\left(\frac{\exp\left({D(\bar{x})}^T
		D(x)/t\right)}{\exp\left({D(\bar{x})}^T D(x)/t\right) + \sum_{j=1}^{M}\exp\left({D(\bar{x})}^T
		D(\tilde{x}_j)/t\right)}\right) \right],$}
\end{equation}
where we build $M$ different corrupted versions of $x$, \textit{i.e.} $\{\tilde{x}_1,\tilde{x}_2,...,\tilde{x}_M\}$, as the negative samples, and $t$ is the temperature that controls the push and pull forces.

Our total contrastive loss $\mathcal{L}_{contra}$ is defined by combining Equation \ref{eq:3} and Equation \ref{eq:4}:
\begin{equation}
\mathcal{L}_{contra} = \lambda_{text}\mathcal{L}_{contra}^{text} +
\lambda_{sem}\mathcal{L}_{contra}^{sem},
\end{equation}
where $\lambda_{text}$ and $\lambda_{sem}$ are the weights that balance the two corresponding losses, respectively.

\subsection{Other Training Objectives}
Aside from our introduced segmentaion confusion adversarial training and contrastive learning
objectives,
we additionaly use two training objectives for optimization. A standard global adversarial training loss
is used in our framework to encourage the global consistency of the repaired images as follows:
\begin{equation}
\mathcal{L}_{adv} = \min_G \max_D \mathbb{E}_{x} [\log D(x)] + \mathbb{E}_{\bar{x}}[\log
(1-D(\bar{x}))].
\end{equation}
We also use an $L$1 reconstruction loss to encourage the inpainting image to be the same as the
ground truth image:
\begin{equation}
\mathcal{L}_{rec} = \mathbb{E}||\hat{x} - x||_1.
\end{equation}

\subsection{Overall Objective}
The overall training objective for our model is defined as follows:
\begin{equation}
\mathcal{L}_{total} = \lambda_{adv} (\mathcal{L}_{adv} +
\mathcal{L}_{\text{SCAT}}) +
\mathcal{L}_{contra} + \lambda_{rec} \mathcal{L}_{rec}
\end{equation}
where $\lambda_{adv}$ and $\lambda_{rec}$ are the weights that control
the importance of the corresponding losses, respectively.

\begin{figure}[t]
	\centering
	\includegraphics[width=1.0\linewidth]{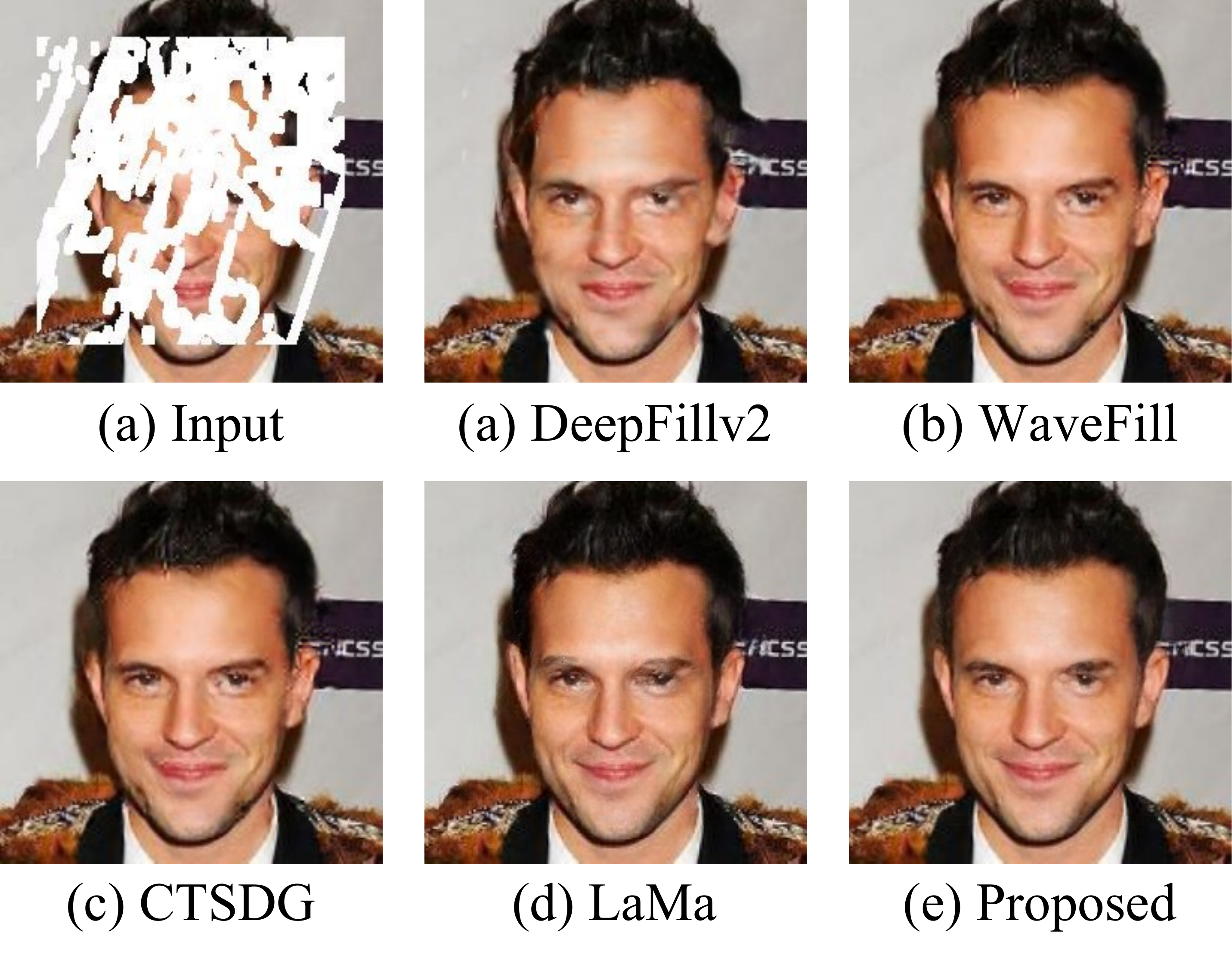}
	\caption{Qualitative results on CelebA. Please zoom in to see more details.}
	\label{fig:celeba}
\end{figure}

\begin{table*}[t]
	\small
	\centering
	\begin{tabular}{c|l|ccc|ccc}
		\hline
		\hline
		\multicolumn{2}{c|}{Dataset} & \multicolumn{3}{|c|}{Places2} &
		\multicolumn{3}{|c}{CelebA} \\
		\hline
		\multicolumn{2}{c|}{Mask Ratio} & 0\%-20\% & 20\%-40\% & 40\%-60\% & 0\%-20\% &
		20\%-40\% & 40\%-60\% \\
		\hline
		\hline

		\multirow{6}{*}{Mean $\mathit{l}_1^{\dag}$}
		&DeepFillv2         & 0.017 & 0.053 & 0.112 & 0.0102 & 0.0316 & 0.0729
		\\
		&WaveFill           & 0.015 & 0.042 & 0.117 & 0.0086 & 0.0230 & 0.0580
		\\
		&CTSDG & 0.012 & 0.041 & 0.089 & 0.0062 & 0.0235 & 0.0567
		\\
		&CoModGAN & 0.015 & 0.049 & 0.109 & N/A & N/A & N/A
		\\
		&LaMa & 0.012 & 0.042 & 0.089 & 0.0070 & 0.0255 & 0.0609
		\\
		&Proposed                 & \textbf{0.010} & \textbf{0.038} & \textbf{0.087} &
		\textbf{0.0057} & \textbf{0.0217} & \textbf{0.0548} \\
		\hline

		\multirow{6}{*}{PSNR$^{\star}$}
		&DeepFillv2        & 30.37 & 22.92 & 18.76 & 35.47 & 27.36 & 22.18 \\
		&WaveFill           & 30.73 & 25.26 & 19.73 & 37.24 & 29.54 & 24.23 \\
		&CTSDG & 32.76 & 24.98 & 20.66 & 38.08 & 29.33 & 24.10  \\
		&CoModGAN & 31.35 & 23.50 & 18.97 & N/A & N/A & N/A  \\
		&LaMa & 32.13 & 24.71 & 20.52 & 37.12 & 28.87 & 23.64  \\
		&Proposed                & \textbf{33.52} & \textbf{25.50} & \textbf{20.85} & \textbf{38.87} &
		\textbf{30.14} & \textbf{24.49} \\
		\hline

		\multirow{6}{*}{SSIM$^{\star}$}
		&DeepFillv2        & 0.946 & 0.829 & 0.680 & 0.968 & 0.896 & 0.783 \\
		&WaveFill         & 0.950 & 0.860 & 0.674 & 0.976 & 0.922 & 0.832 \\
		&CTSDG & 0.959 & 0.857 & 0.716 & 0.979 & 0.920 & 0.824 \\
		&CoModGAN & 0.953 & 0.842 & 0.690 & N/A & N/A & N/A \\
		&LaMa & 0.956 & 0.856 & 0.721 & 0.974 & 0.910 & 0.810 \\
		&Proposed                & \textbf{0.964} & \textbf{0.870} & \textbf{0.732} & \textbf{0.981} &
		\textbf{0.928} & \textbf{0.836}  \\
		\hline

		\multirow{6}{*}{FID$^{\dag}$}
		&DeepFillv2          & 3.07 & 10.48 & 24.35 & 1.82 & 5.19 & 10.25 \\
		&WaveFill         & 3.23 & 7.03 & 39.35 & 1.27 & 2.45 & 7.57 \\
		&CTSDG & 2.74 & 11.27 & 31.45 & 1.09 & 4.72 & 12.26 \\
		&CoModGAN & 2.00 & 6.29 & 15.51 & N/A & N/A & N/A \\
		&LaMa & 1.85 & 5.82 & 14.68 & 1.07 & 3.51 & 7.66 \\
		&Proposed                 & \textbf{1.68} & \textbf{5.71} & \textbf{13.83} &
		\textbf{0.71} & \textbf{2.40} & \textbf{6.15} \\
		\hline
		\hline
	\end{tabular}
	\caption{Quantitative comparison on Places2 and CelebA datasets. $^{\star}$Higher is better.
		$^{\dag}$Lower is
		better. N/A indicates the result is not available.}
	\label{tab:quantitative_comparison}
\end{table*}

\begin{figure*}[t!]
	\centering
	\includegraphics[width=\linewidth]{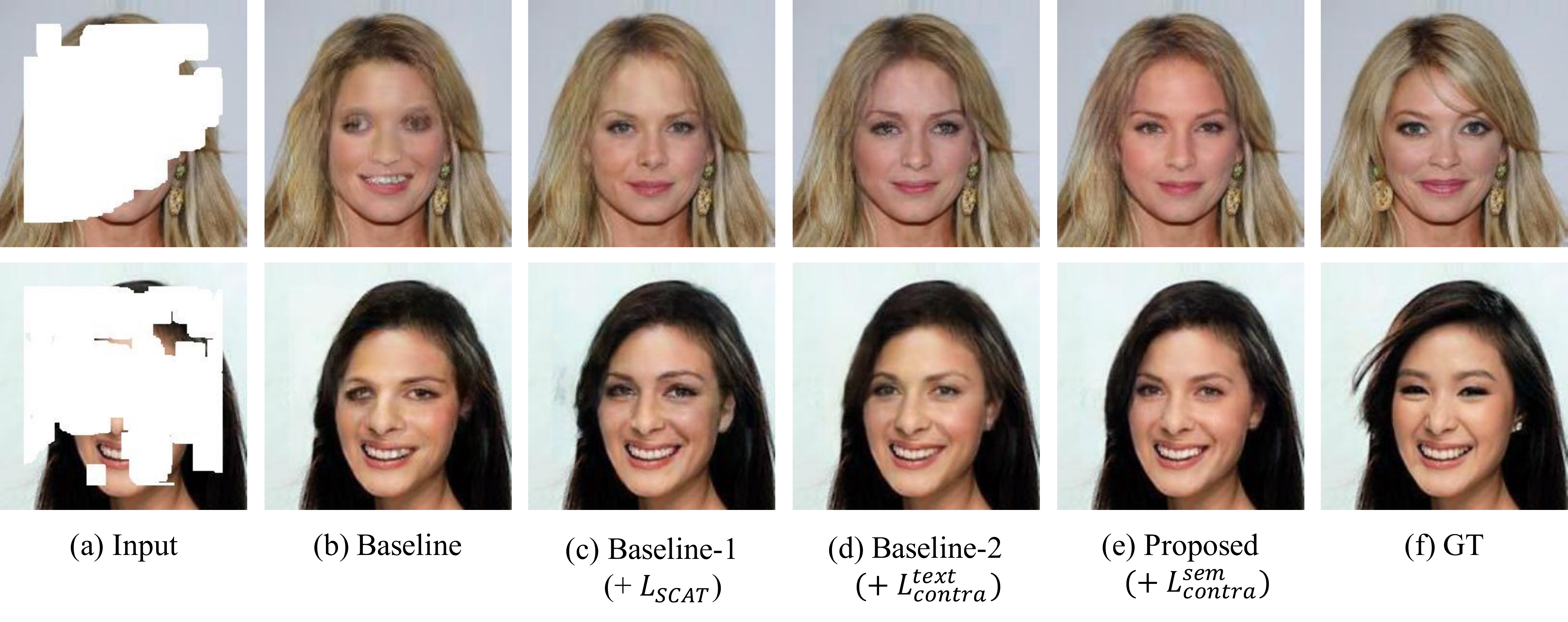}
	\caption{Qualitative results of ablation studies on CelebA. Please zoom in to see more details.}
	\label{fig:celeba_ablation}
\end{figure*}

\section{Experiments}
\subsection{Experimental Settings}
\label{subsec:details}
\paragraph{Implementation Details}
We train our model with a batch size of 8 on a single 24G NVIDIA RTX3090 GPU. 
All the masks and images for training and evaluation are of size $256\times256$. Our inpainting
generator consists of a few downsampling and upsampling layers with several AOT-blocks~\cite{zeng2022aggregated} in-between. The segmentation network $S$ in our framework is
implemented as a U-net. We adopt hinge loss~\cite{lim2017geometric} for both global adversarial training and SCAT and apply spectral normalization~\cite{miyato2018spectral} on all $S$
and $D$'s layers. The negative sample size for the semantic contrastive learning loss is $8$. We conduct experiments on the CelebA dataset to select the hyper-parameters from a set of empirical values, i.e. $[0.1, 1, 5, 10]$, and find that setting $\lambda_{adv}=1$,
$\lambda_{text}=10$, $\lambda_{sem}=1$, and
$\lambda_{rec}=10$ works fine for our model.

\begin{table*}[t]
	\centering
		\begin{tabular}{c|c|c|c|c}
			\hline
			\hline
			\multirow{2}{*}{Models} & Baseline &
			Baseline-1  & Baseline-2  & Proposed \\
			& (w/o $\mathcal{L}_{\text{SCAT}}$ \& $\mathcal{L}_{contra}$) & (+
			$\mathcal{L}_{\text{SCAT}}$) & (+ $\mathcal{L}_{contra}^{text}$) & (+
			$\mathcal{L}_{contra}^{sem}$)\\
			\hline
			
			Mean $\mathit{l}_1^{\dag}$ & 0.0593 & 0.0576 & 0.0561 & \textbf{0.0548} \\
			\hline
			PSNR$^{\star}$ & 23.86 & 24.11 & 24.27 & \textbf{24.49} \\
			
			\hline
			
			SSIM$^{\star}$ & 0.822 & 0.825 & 0.832 & \textbf{0.836}\\
			
			\hline
			
			FID$^{\dag}$ & 7.84 & 7.20 & 6.52 & \textbf{6.15}\\
			\hline
			\hline
		\end{tabular}
	\caption{Quantitative results of ablation studies on CelebA. The results are obtained using
		masks with mask ratio from 40\% to 60\%. $^{\star}$Higher is better. $^{\dag}$Lower is better.}
	\label{tab:celeba_ablation}
\end{table*}

\paragraph{Datasets.}
We adopt an irregular mask dataset~\cite{liu2018image} for training and evaluation, which contains 12,000 masks with mask ratio from 0\% to 60\%. We train and evaluate our method on two benchmark datasets Places2~\cite{zhou2017places} and CelebA~\cite{liu2018large} following their official training/validation splits.

\paragraph{Baselines.}
We compare our method with five state-of-the-art methods: DeepFillv2~\cite{yu2019free},
CTSDG~\cite{guo2021image}, WaveFill~\cite{yu2021wavefill}, CoModGAN~\cite{zhao2021large},
and
LaMa~\cite{suvorov2022resolution} by using their
officially-released pre-trained models or models trained
with their officially-released codes.


\paragraph{Metrics.}
Four widely-adopted quantitative metrics are used for evaluation:	mean $\mathit{l}_1$ error,
PSNR, SSIM, and FID~\cite{heusel2017gans}.

\subsection{Qualitative Evaluations}
Figure \ref{fig:places2} and \ref{fig:celeba}  compare our method's inpainting results with the
state-of-the-art methods' on Places2 and CelebA, respectively. As shown in Figure \ref{fig:places2}, thanks to our dual adversarial training in both global and pixel level plus contrastive losses,
our method performs more favorably comprared to the other state-of-the-art methods in maintaining
the structural integrity of the objects and repairing large-scale missing areas in the natural landscape.
LaMa and CoModGAN sometimes produce inpainting results with distorted structures. CTSDG tends to
create small edge-like artifacts, which may be casued by the incorrect edge predictions in its
framework. WaveFill is prone to synthesize blurry contents in the context of large-scale missing
textures and DeepFillv2's results often contain distorted structures or unrealistic artifacts. For results on CelebA,
our method's results are comparable to the state-of-the-art methods, exhibiting global consistency and
fewer local artifacts (\textit{e.g.,} the areas of eyes and ears in Figure \ref{fig:celeba}). The qualitative
comparison on the two datasets subjectively verify the superiority of our method.

\subsection{Quantitative Evaluations}
We perform quantitative evaluations on Places2 and CelebA datasets by randomly selecting
5K images from each dataset's validation set and then applying the same irregular masks on them to
obtain the quantitative results. Table \ref{tab:quantitative_comparison} presents
the numerical results with different mask ratios on the two datsets. As we see, the proposed
method outperforms the other state-of-the-art methods in all cases by considerable margins,
especially on FID, objectively demonstrating the superiority of our method. \\

\subsection{Ablation Studies}
We perform ablation studies of our proposed methods on CelebA dataset to evaluate their
effectiveness. We denote the model that ablates all our proposed methods, including the SCAT and the
two contrastive learning objectives, as the Baseline model. The Baseline model actually follows the
global adversarial training framework of CE~\cite{pathak2016context}. We then augment the Baseline
model with the SCAT $\mathcal{L}_{\text{SCAT}}$, the textural contrastive learning
$\mathcal{L}_{contra}^{text}$, and the semantic contrastive learning $\mathcal{L}_{contra}^{sem}$
sequentially for training, where the resulting models are denoted as Baseline-1, Baseline-2, and the
Proposed full model, accordingly.  We show the qualitative and  quantitative results of the ablation
studies in Figure \ref{fig:celeba_ablation} and Table \ref{tab:celeba_ablation}, respectively. As can be
observed from Figure \ref{fig:celeba_ablation}, the Baseline model creates globally consistent
completed images, but the results contain obvious local artifacts (\textit{e.g.}, the eyes). By adding the
SCAT, Baseline-1 produces inpainting images with finer local details. Further, by adding the textural
contrastive learning, Baseline-2 makes the textural details of the repaired images more realistic
(\textit{e.g.,} smoother skins and less artifacts around the eyes, the ears, and hairs). Finally, the
Proposed full model's inpainting results are more visually appealing and globally coherent by adding the
semantic contrastive learning. Also, as shown in Table \ref{tab:celeba_ablation}, each of our proposed
losses helps improve the quantitative metrics, and the Proposed full model achieves the best results.
Both the qualitative and quantitative results of our ablation studies verify the proposed methods'
effectiveness.

\section{Concluding Remarks}
A new adversarial training framework for image
inpainting with segmentation confusion adversarial training (SCAT) and
contrastive learning is presented in the paper. SCAT formulates a novel adversarial game
between an inpainting generator and a segmentation network,
providing pixel-level local training signals and the flexibility of
processing images with free-form holes for our framework. On the other hand,
the proposed contrastive learning losses stabilize and improve our
framework’s training and bring more realistic inpainting images by pulling the
inpainting images closer to the ground truth images and pushing them farther
away from the corrupted images in the representation space of
the discriminator. Extensive experiments on two benchmark datasets
demonstrate the effectiveness of the proposed method. Future work may include verifying the effectiveness of the proposed objectives on other GAN-based inpainting frameworks and extending our framework to diversified inpainting.

\section{Acknowledgments}
This work was supported in part by the National Key Research and Development Program of China (2020YFC1523101), the National Natural Science Foundation of China (62172365), Zhejiang Elite Program (2022C01222), the Key Program of the National Social Science Foundation of China (19ZDA197), and MOE Frontier Science Center for Brain Science\& Brain-Machine Integration (Zhejiang University).

\bibliography{aaai23}

\begin{thebibliography}{60}
\providecommand{\natexlab}[1]{#1}

\bibitem[{Arjovsky, Chintala, and Bottou(2017)}]{arjovsky2017wasserstein}
Arjovsky, M.; Chintala, S.; and Bottou, L. 2017.
\newblock Wasserstein gan.
\newblock \emph{arXiv preprint arXiv:1701.07875}.

\bibitem[{Barnes et~al.(2009)Barnes, Shechtman, Finkelstein, and
  Goldman}]{barnes2009patchmatch}
Barnes, C.; Shechtman, E.; Finkelstein, A.; and Goldman, D.~B. 2009.
\newblock PatchMatch: A randomized correspondence algorithm for structural
  image editing.
\newblock \emph{ACM Trans. Graph.}, 28(3): 24.

\bibitem[{Bertalmio et~al.(2000)Bertalmio, Sapiro, Caselles, and
  Ballester}]{bertalmio2000image}
Bertalmio, M.; Sapiro, G.; Caselles, V.; and Ballester, C. 2000.
\newblock Image inpainting.
\newblock In \emph{Proceedings of the 27th annual conference on Computer
  graphics and interactive techniques}, 417--424.

\bibitem[{Bertalmio et~al.(2003)Bertalmio, Vese, Sapiro, and
  Osher}]{bertalmio2003simultaneous}
Bertalmio, M.; Vese, L.; Sapiro, G.; and Osher, S. 2003.
\newblock Simultaneous structure and texture image inpainting.
\newblock \emph{IEEE transactions on image processing}, 12(8): 882--889.

\bibitem[{Brock, Donahue, and Simonyan(2018)}]{brock2018large}
Brock, A.; Donahue, J.; and Simonyan, K. 2018.
\newblock Large scale GAN training for high fidelity natural image synthesis.
\newblock \emph{arXiv preprint arXiv:1809.11096}.

\bibitem[{Chen et~al.(2021)Chen, Wang, Zhang, Zuo, Li, Xing, Lu
  et~al.}]{chen2021artistic}
Chen, H.; Wang, Z.; Zhang, H.; Zuo, Z.; Li, A.; Xing, W.; Lu, D.; et~al. 2021.
\newblock Artistic Style Transfer with Internal-external Learning and
  Contrastive Learning.
\newblock \emph{Advances in Neural Information Processing Systems}, 34.

\bibitem[{Chen et~al.(2020)Chen, Kornblith, Norouzi, and
  Hinton}]{chen2020simple}
Chen, T.; Kornblith, S.; Norouzi, M.; and Hinton, G. 2020.
\newblock A simple framework for contrastive learning of visual
  representations.
\newblock In \emph{International conference on machine learning}, 1597--1607.
  PMLR.

\bibitem[{Criminisi, P{\'e}rez, and Toyama(2004)}]{criminisi2004region}
Criminisi, A.; P{\'e}rez, P.; and Toyama, K. 2004.
\newblock Region filling and object removal by exemplar-based image inpainting.
\newblock \emph{IEEE Transactions on image processing}, 13(9): 1200--1212.

\bibitem[{Donahue, Kr{\"a}henb{\"u}hl, and
  Darrell(2016)}]{donahue2016adversarial}
Donahue, J.; Kr{\"a}henb{\"u}hl, P.; and Darrell, T. 2016.
\newblock Adversarial feature learning.
\newblock \emph{arXiv preprint arXiv:1605.09782}.

\bibitem[{Dong, Cao, and Fu(2022)}]{dong2022incremental}
Dong, Q.; Cao, C.; and Fu, Y. 2022.
\newblock Incremental transformer structure enhanced image inpainting with
  masking positional encoding.
\newblock In \emph{Proceedings of the IEEE/CVF Conference on Computer Vision
  and Pattern Recognition}, 11358--11368.

\bibitem[{Dumoulin et~al.(2016)Dumoulin, Belghazi, Poole, Mastropietro, Lamb,
  Arjovsky, and Courville}]{dumoulin2016adversarially}
Dumoulin, V.; Belghazi, I.; Poole, B.; Mastropietro, O.; Lamb, A.; Arjovsky,
  M.; and Courville, A. 2016.
\newblock Adversarially learned inference.
\newblock \emph{arXiv preprint arXiv:1606.00704}.

\bibitem[{Goodfellow(2016)}]{goodfellow2016nips}
Goodfellow, I. 2016.
\newblock NIPS 2016 tutorial: Generative adversarial networks.
\newblock \emph{arXiv preprint arXiv:1701.00160}.

\bibitem[{Goodfellow et~al.(2014)Goodfellow, Pouget-Abadie, Mirza, Xu,
  Warde-Farley, Ozair, Courville, and Bengio}]{goodfellow2014generative}
Goodfellow, I.; Pouget-Abadie, J.; Mirza, M.; Xu, B.; Warde-Farley, D.; Ozair,
  S.; Courville, A.; and Bengio, Y. 2014.
\newblock Generative adversarial nets.
\newblock \emph{Advances in neural information processing systems}, 27.

\bibitem[{Gulrajani et~al.(2017)Gulrajani, Ahmed, Arjovsky, Dumoulin, and
  Courville}]{gulrajani2017improved}
Gulrajani, I.; Ahmed, F.; Arjovsky, M.; Dumoulin, V.; and Courville, A.~C.
  2017.
\newblock Improved training of wasserstein gans.
\newblock In \emph{Advances in neural information processing systems},
  5767--5777.

\bibitem[{Guo, Yang, and Huang(2021)}]{guo2021image}
Guo, X.; Yang, H.; and Huang, D. 2021.
\newblock Image Inpainting via Conditional Texture and Structure Dual
  Generation.
\newblock In \emph{Proceedings of the IEEE/CVF International Conference on
  Computer Vision}, 14134--14143.

\bibitem[{Guo et~al.(2019)Guo, Chen, Yu, Chen, and Liu}]{guo2019progressive}
Guo, Z.; Chen, Z.; Yu, T.; Chen, J.; and Liu, S. 2019.
\newblock Progressive image inpainting with full-resolution residual network.
\newblock In \emph{Proceedings of the 27th acm international conference on
  multimedia}, 2496--2504.

\bibitem[{He et~al.(2020)He, Fan, Wu, Xie, and Girshick}]{he2020momentum}
He, K.; Fan, H.; Wu, Y.; Xie, S.; and Girshick, R. 2020.
\newblock Momentum contrast for unsupervised visual representation learning.
\newblock In \emph{Proceedings of the IEEE/CVF conference on computer vision
  and pattern recognition}, 9729--9738.

\bibitem[{He et~al.(2016)He, Zhang, Ren, and Sun}]{he2016deep}
He, K.; Zhang, X.; Ren, S.; and Sun, J. 2016.
\newblock Deep residual learning for image recognition.
\newblock In \emph{Proceedings of the IEEE conference on computer vision and
  pattern recognition}, 770--778.

\bibitem[{Heusel et~al.(2017)Heusel, Ramsauer, Unterthiner, Nessler, and
  Hochreiter}]{heusel2017gans}
Heusel, M.; Ramsauer, H.; Unterthiner, T.; Nessler, B.; and Hochreiter, S.
  2017.
\newblock Gans trained by a two time-scale update rule converge to a local nash
  equilibrium.
\newblock In \emph{Advances in neural information processing systems},
  6626--6637.

\bibitem[{Huang et~al.(2014)Huang, Kang, Ahuja, and Kopf}]{huang2014image}
Huang, J.-B.; Kang, S.~B.; Ahuja, N.; and Kopf, J. 2014.
\newblock Image completion using planar structure guidance.
\newblock \emph{ACM Transactions on graphics (TOG)}, 33(4): 1--10.

\bibitem[{Iizuka, Simo-Serra, and Ishikawa(2017)}]{iizuka2017globally}
Iizuka, S.; Simo-Serra, E.; and Ishikawa, H. 2017.
\newblock Globally and locally consistent image completion.
\newblock \emph{ACM Transactions on Graphics (ToG)}, 36(4): 1--14.

\bibitem[{Isola et~al.(2017)Isola, Zhu, Zhou, and Efros}]{isola2017image}
Isola, P.; Zhu, J.-Y.; Zhou, T.; and Efros, A.~A. 2017.
\newblock Image-to-image translation with conditional adversarial networks.
\newblock In \emph{Proceedings of the IEEE conference on computer vision and
  pattern recognition}, 1125--1134.

\bibitem[{Kang and Park(2020)}]{kang2020contragan}
Kang, M.; and Park, J. 2020.
\newblock Contragan: Contrastive learning for conditional image generation.
\newblock \emph{Advances in Neural Information Processing Systems}, 33:
  21357--21369.

\bibitem[{Karras, Laine, and Aila(2019)}]{karras2019style}
Karras, T.; Laine, S.; and Aila, T. 2019.
\newblock A style-based generator architecture for generative adversarial
  networks.
\newblock In \emph{Proceedings of the IEEE conference on computer vision and
  pattern recognition}, 4401--4410.

\bibitem[{Krizhevsky, Sutskever, and Hinton(2012)}]{krizhevsky2012imagenet}
Krizhevsky, A.; Sutskever, I.; and Hinton, G.~E. 2012.
\newblock Imagenet classification with deep convolutional neural networks.
\newblock \emph{Advances in neural information processing systems}, 25.

\bibitem[{Larsen et~al.(2016)Larsen, S{\o}nderby, Larochelle, and
  Winther}]{larsen2016autoencoding}
Larsen, A. B.~L.; S{\o}nderby, S.~K.; Larochelle, H.; and Winther, O. 2016.
\newblock Autoencoding beyond pixels using a learned similarity metric.
\newblock In \emph{International conference on machine learning}, 1558--1566.

\bibitem[{Li et~al.(2020)Li, Wang, Zhang, Du, and Tao}]{li2020recurrent}
Li, J.; Wang, N.; Zhang, L.; Du, B.; and Tao, D. 2020.
\newblock Recurrent feature reasoning for image inpainting.
\newblock In \emph{Proceedings of the IEEE/CVF Conference on Computer Vision
  and Pattern Recognition}, 7760--7768.

\bibitem[{Li et~al.(2022)Li, Lin, Zhou, Qi, Wang, and Jia}]{li2022mat}
Li, W.; Lin, Z.; Zhou, K.; Qi, L.; Wang, Y.; and Jia, J. 2022.
\newblock MAT: Mask-Aware Transformer for Large Hole Image Inpainting.
\newblock In \emph{Proceedings of the IEEE/CVF Conference on Computer Vision
  and Pattern Recognition}, 10758--10768.

\bibitem[{Lim and Ye(2017)}]{lim2017geometric}
Lim, J.~H.; and Ye, J.~C. 2017.
\newblock Geometric gan.
\newblock \emph{arXiv preprint arXiv:1705.02894}.

\bibitem[{Liu et~al.(2018{\natexlab{a}})Liu, Reda, Shih, Wang, Tao, and
  Catanzaro}]{liu2018image}
Liu, G.; Reda, F.~A.; Shih, K.~J.; Wang, T.-C.; Tao, A.; and Catanzaro, B.
  2018{\natexlab{a}}.
\newblock Image inpainting for irregular holes using partial convolutions.
\newblock In \emph{Proceedings of the European conference on computer vision
  (ECCV)}, 85--100.

\bibitem[{Liu et~al.(2022)Liu, Tan, Chen, Chu, Dai, Chen, Liu, Yuan, and
  Yu}]{liu2022reduce}
Liu, Q.; Tan, Z.; Chen, D.; Chu, Q.; Dai, X.; Chen, Y.; Liu, M.; Yuan, L.; and
  Yu, N. 2022.
\newblock Reduce Information Loss in Transformers for Pluralistic Image
  Inpainting.
\newblock In \emph{Proceedings of the IEEE/CVF Conference on Computer Vision
  and Pattern Recognition}, 11347--11357.

\bibitem[{Liu et~al.(2018{\natexlab{b}})Liu, Luo, Wang, and
  Tang}]{liu2018large}
Liu, Z.; Luo, P.; Wang, X.; and Tang, X. 2018{\natexlab{b}}.
\newblock Large-scale celebfaces attributes (celeba) dataset.
\newblock \emph{Retrieved August}, 15(2018): 11.

\bibitem[{Ma et~al.(2020)Ma, Zhou, Huang, Chai, Wei, and He}]{ma2020can}
Ma, X.; Zhou, X.; Huang, H.; Chai, Z.; Wei, X.; and He, R. 2020.
\newblock Free-form image inpainting via contrastive attention network.
\newblock \emph{arXiv preprint arXiv:2010.15643}.

\bibitem[{Miyato et~al.(2018)Miyato, Kataoka, Koyama, and
  Yoshida}]{miyato2018spectral}
Miyato, T.; Kataoka, T.; Koyama, M.; and Yoshida, Y. 2018.
\newblock Spectral normalization for generative adversarial networks.
\newblock \emph{arXiv preprint arXiv:1802.05957}.

\bibitem[{Nazeri et~al.(2019)Nazeri, Ng, Joseph, Qureshi, and
  Ebrahimi}]{nazeri2019edgeconnect}
Nazeri, K.; Ng, E.; Joseph, T.; Qureshi, F.~Z.; and Ebrahimi, M. 2019.
\newblock Edgeconnect: Generative image inpainting with adversarial edge
  learning.
\newblock \emph{arXiv preprint arXiv:1901.00212}.

\bibitem[{Park et~al.(2020)Park, Efros, Zhang, and Zhu}]{park2020contrastive}
Park, T.; Efros, A.~A.; Zhang, R.; and Zhu, J.-Y. 2020.
\newblock Contrastive learning for unpaired image-to-image translation.
\newblock In \emph{European Conference on Computer Vision}, 319--345. Springer.

\bibitem[{Pathak et~al.(2016)Pathak, Krahenbuhl, Donahue, Darrell, and
  Efros}]{pathak2016context}
Pathak, D.; Krahenbuhl, P.; Donahue, J.; Darrell, T.; and Efros, A.~A. 2016.
\newblock Context encoders: Feature learning by inpainting.
\newblock In \emph{Proceedings of the IEEE conference on computer vision and
  pattern recognition}, 2536--2544.

\bibitem[{Radford, Metz, and Chintala(2015)}]{radford2015unsupervised}
Radford, A.; Metz, L.; and Chintala, S. 2015.
\newblock Unsupervised representation learning with deep convolutional
  generative adversarial networks.
\newblock \emph{arXiv preprint arXiv:1511.06434}.

\bibitem[{Salimans et~al.(2016)Salimans, Goodfellow, Zaremba, Cheung, Radford,
  and Chen}]{salimans2016improved}
Salimans, T.; Goodfellow, I.; Zaremba, W.; Cheung, V.; Radford, A.; and Chen,
  X. 2016.
\newblock Improved techniques for training gans.
\newblock In \emph{Advances in neural information processing systems},
  2234--2242.

\bibitem[{Schonfeld, Schiele, and Khoreva(2020)}]{schonfeld2020u}
Schonfeld, E.; Schiele, B.; and Khoreva, A. 2020.
\newblock A u-net based discriminator for generative adversarial networks.
\newblock In \emph{Proceedings of the IEEE/CVF Conference on Computer Vision
  and Pattern Recognition}, 8207--8216.

\bibitem[{Simonyan and Zisserman(2014)}]{simonyan2014very}
Simonyan, K.; and Zisserman, A. 2014.
\newblock Very deep convolutional networks for large-scale image recognition.
\newblock \emph{arXiv preprint arXiv:1409.1556}.

\bibitem[{Srivastava et~al.(2017)Srivastava, Valkov, Russell, Gutmann, and
  Sutton}]{srivastava2017veegan}
Srivastava, A.; Valkov, L.; Russell, C.; Gutmann, M.~U.; and Sutton, C. 2017.
\newblock Veegan: Reducing mode collapse in gans using implicit variational
  learning.
\newblock In \emph{Advances in Neural Information Processing Systems},
  3308--3318.

\bibitem[{Suvorov et~al.(2022)Suvorov, Logacheva, Mashikhin, Remizova, Ashukha,
  Silvestrov, Kong, Goka, Park, and Lempitsky}]{suvorov2022resolution}
Suvorov, R.; Logacheva, E.; Mashikhin, A.; Remizova, A.; Ashukha, A.;
  Silvestrov, A.; Kong, N.; Goka, H.; Park, K.; and Lempitsky, V. 2022.
\newblock Resolution-robust Large Mask Inpainting with Fourier Convolutions.
\newblock In \emph{Proceedings of the IEEE/CVF Winter Conference on
  Applications of Computer Vision}, 2149--2159.

\bibitem[{Szegedy et~al.(2015)Szegedy, Liu, Jia, Sermanet, Reed, Anguelov,
  Erhan, Vanhoucke, and Rabinovich}]{szegedy2015going}
Szegedy, C.; Liu, W.; Jia, Y.; Sermanet, P.; Reed, S.; Anguelov, D.; Erhan, D.;
  Vanhoucke, V.; and Rabinovich, A. 2015.
\newblock Going deeper with convolutions.
\newblock In \emph{Proceedings of the IEEE conference on computer vision and
  pattern recognition}, 1--9.

\bibitem[{Thanh-Tung and Tran(2020)}]{thanh2020catastrophic}
Thanh-Tung, H.; and Tran, T. 2020.
\newblock Catastrophic forgetting and mode collapse in gans.
\newblock In \emph{2020 international joint conference on neural networks
  (ijcnn)}, 1--10. IEEE.

\bibitem[{Ulyanov, Vedaldi, and Lempitsky(2018)}]{ulyanov2018takes}
Ulyanov, D.; Vedaldi, A.; and Lempitsky, V. 2018.
\newblock It takes (only) two: Adversarial generator-encoder networks.
\newblock In \emph{Thirty-Second AAAI Conference on Artificial Intelligence}.

\bibitem[{Wan et~al.(2021)Wan, Zhang, Chen, and Liao}]{wan2021high}
Wan, Z.; Zhang, J.; Chen, D.; and Liao, J. 2021.
\newblock High-fidelity pluralistic image completion with transformers.
\newblock In \emph{Proceedings of the IEEE/CVF International Conference on
  Computer Vision}, 4692--4701.

\bibitem[{Wu et~al.(2021)Wu, Qu, Lin, Zhou, Qiao, Zhang, Xie, and
  Ma}]{wu2021contrastive}
Wu, H.; Qu, Y.; Lin, S.; Zhou, J.; Qiao, R.; Zhang, Z.; Xie, Y.; and Ma, L.
  2021.
\newblock Contrastive learning for compact single image dehazing.
\newblock In \emph{Proceedings of the IEEE/CVF Conference on Computer Vision
  and Pattern Recognition}, 10551--10560.

\bibitem[{Yu et~al.(2018)Yu, Lin, Yang, Shen, Lu, and Huang}]{yu2018generative}
Yu, J.; Lin, Z.; Yang, J.; Shen, X.; Lu, X.; and Huang, T.~S. 2018.
\newblock Generative image inpainting with contextual attention.
\newblock In \emph{Proceedings of the IEEE conference on computer vision and
  pattern recognition}, 5505--5514.

\bibitem[{Yu et~al.(2019)Yu, Lin, Yang, Shen, Lu, and Huang}]{yu2019free}
Yu, J.; Lin, Z.; Yang, J.; Shen, X.; Lu, X.; and Huang, T.~S. 2019.
\newblock Free-form image inpainting with gated convolution.
\newblock In \emph{Proceedings of the IEEE/CVF International Conference on
  Computer Vision}, 4471--4480.

\bibitem[{Yu et~al.(2021)Yu, Zhan, Lu, Pan, Ma, Xie, and Miao}]{yu2021wavefill}
Yu, Y.; Zhan, F.; Lu, S.; Pan, J.; Ma, F.; Xie, X.; and Miao, C. 2021.
\newblock WaveFill: A Wavelet-based Generation Network for Image Inpainting.
\newblock In \emph{Proceedings of the IEEE/CVF International Conference on
  Computer Vision}, 14114--14123.

\bibitem[{Zeiler and Fergus(2014)}]{zeiler2014visualizing}
Zeiler, M.~D.; and Fergus, R. 2014.
\newblock Visualizing and understanding convolutional networks.
\newblock In \emph{European conference on computer vision}, 818--833. Springer.

\bibitem[{Zeng et~al.(2019)Zeng, Fu, Chao, and Guo}]{zeng2019learning}
Zeng, Y.; Fu, J.; Chao, H.; and Guo, B. 2019.
\newblock Learning pyramid-context encoder network for high-quality image
  inpainting.
\newblock In \emph{Proceedings of the IEEE/CVF Conference on Computer Vision
  and Pattern Recognition}, 1486--1494.

\bibitem[{Zeng et~al.(2022)Zeng, Fu, Chao, and Guo}]{zeng2022aggregated}
Zeng, Y.; Fu, J.; Chao, H.; and Guo, B. 2022.
\newblock Aggregated contextual transformations for high-resolution image
  inpainting.
\newblock \emph{IEEE Transactions on Visualization and Computer Graphics}.

\bibitem[{Zeng et~al.(2020)Zeng, Lin, Yang, Zhang, Shechtman, and
  Lu}]{zeng2020high}
Zeng, Y.; Lin, Z.; Yang, J.; Zhang, J.; Shechtman, E.; and Lu, H. 2020.
\newblock High-resolution image inpainting with iterative confidence feedback
  and guided upsampling.
\newblock In \emph{European conference on computer vision}, 1--17. Springer.

\bibitem[{Zhang et~al.(2022)Zhang, Zhou, Barnes, Amirghodsi, Lin, Shechtman,
  and Shi}]{zhang2022perceptual}
Zhang, L.; Zhou, Y.; Barnes, C.; Amirghodsi, S.; Lin, Z.; Shechtman, E.; and
  Shi, J. 2022.
\newblock Perceptual artifacts localization for inpainting.
\newblock In \emph{Computer Vision--ECCV 2022: 17th European Conference, Tel
  Aviv, Israel, October 23--27, 2022, Proceedings, Part XXIX}, 146--164.
  Springer.

\bibitem[{Zhao et~al.(2020)Zhao, Mo, Lin, Wang, Zuo, Chen, Xing, and
  Lu}]{zhao2020uctgan}
Zhao, L.; Mo, Q.; Lin, S.; Wang, Z.; Zuo, Z.; Chen, H.; Xing, W.; and Lu, D.
  2020.
\newblock Uctgan: Diverse image inpainting based on unsupervised cross-space
  translation.
\newblock In \emph{Proceedings of the IEEE/CVF conference on computer vision
  and pattern recognition}, 5741--5750.

\bibitem[{Zhao et~al.(2021)Zhao, Cui, Sheng, Dong, Liang, Chang, and
  Xu}]{zhao2021large}
Zhao, S.; Cui, J.; Sheng, Y.; Dong, Y.; Liang, X.; Chang, E.~I.; and Xu, Y.
  2021.
\newblock Large scale image completion via co-modulated generative adversarial
  networks.
\newblock \emph{arXiv preprint arXiv:2103.10428}.

\bibitem[{Zheng, Cham, and Cai(2019)}]{zheng2019pluralistic}
Zheng, C.; Cham, T.-J.; and Cai, J. 2019.
\newblock Pluralistic image completion.
\newblock In \emph{Proceedings of the IEEE/CVF Conference on Computer Vision
  and Pattern Recognition}, 1438--1447.

\bibitem[{Zhou et~al.(2017)Zhou, Lapedriza, Khosla, Oliva, and
  Torralba}]{zhou2017places}
Zhou, B.; Lapedriza, A.; Khosla, A.; Oliva, A.; and Torralba, A. 2017.
\newblock Places: A 10 million image database for scene recognition.
\newblock \emph{IEEE transactions on pattern analysis and machine
  intelligence}, 40(6): 1452--1464.

\end{thebibliography}

\end{document}